\documentclass{ieeeaccess}
\usepackage{cite}
\usepackage{amsmath,amssymb,amsfonts}
\usepackage{algorithmic}
\usepackage{graphicx}
\graphicspath{{./images/}{./bio}}
\usepackage{booktabs}       % professional-quality tables
\usepackage{multirow}
\usepackage{color}
\usepackage{textcomp}
\def\BibTeX{{\rm B\kern-.05em{\sc i\kern-.025em b}\kern-.08em
    T\kern-.1667em\lower.7ex\hbox{E}\kern-.125emX}}

\begin{document}
\history{Date of publication xxxx 00, 0000, date of current version xxxx 00, 0000.}
\doi{10.1109/ACCESS.2023.0322000}

\title{Blockchain-Enabled Federated Learning: A Reference Architecture Design, Implementation, and Verification}
%Blockchain-Enabled Federated Learning: A Reference Architecture Incorporating a DID Access System}
\author{\uppercase{Eunsu Goh}\authorrefmark{1},
\uppercase{Dae-Yeol Kim}\authorrefmark{1},
\uppercase{Kwangkee Lee}\authorrefmark{1}, 
\uppercase{Suyeong Oh}\authorrefmark{1}, 
\uppercase{Jong-Eui Chae}\authorrefmark{1}, 
\uppercase{Do-Yup Kim}\authorrefmark{2}, \IEEEmembership{Member, IEEE}}

\address[1]{Innopia Technologies Inc., Seongnam-si 13217, South Korea}
\address[2]{Department of Information and Communication AI Engineering, Kyungnam University, Changwon-si, Gyeongsangnam-do 51767, South Korea}
\tfootnote{This work was supported in part by the Commercializations Promotion Agency for Research and Development Outcome (COMPA) Grant funded by the Korean Government (MSIT) through the Future Research Service Development Support under Grant 2022-1-SB1-1.}

\markboth
{E. Goh et al.: Blockchain-Enabled Federated Learning: A Reference Architecture Design, Implementation, and Verification}
{E. Goh et al.: Blockchain-Enabled Federated Learning: A Reference Architecture Design, Implementation, and Verification}

\corresp{Corresponding author: Do-Yup Kim (doyup09@kyungnam.ac.kr) and Kwangkee Lee (kwangkeelee@gmail.com)}

\begin{abstract}
This paper presents a novel reference architecture for blockchain-enabled federated learning (BCFL), a state-of-the-art approach that amalgamates the strengths of federated learning and blockchain technology.
We define smart contract functions, stakeholders and their roles, and the use of interplanetary file system (IPFS) as key components of BCFL and conduct a comprehensive analysis.
In traditional centralized federated learning, the selection of local nodes and the collection of learning results for each round are merged under the control of a central server.
In contrast, in BCFL, all these processes are monitored and managed via smart contracts.
Additionally, we propose an extension architecture to support both cross-device and cross-silo federated learning scenarios.
Furthermore, we implement and verify the architecture in a practical real-world Ethereum development environment.
Our BCFL reference architecture provides significant flexibility and extensibility, accommodating the integration of various additional elements, as per specific requirements and use cases, thereby rendering it an adaptable solution for a wide range of BCFL applications.
As a prominent example of extensibility, decentralized identifiers (DIDs) have been employed as an authentication method to introduce practical utilization within BCFL.
This study not only bridges a crucial gap between research and practical deployment but also lays a solid foundation for future explorations in the realm of BCFL.
The pivotal contribution of this study is the successful implementation and verification of a realistic BCFL reference architecture.
We intend to make the source code publicly accessible shortly, fostering further advancements and adaptations within the community.
\end{abstract}

\begin{keywords}
Blockchain, federated learning, blockchain-enabled federated learning (BCFL), reference architecture, Ethereum test network deployment, decentralized identifier (DID), client selection, client evaluation, smart contracts, data privacy and security.
\end{keywords}

\titlepgskip=-21pt

\maketitle

\section{Introduction}
\label{sec:introduction}
\PARstart{T}{he} field of machine learning is often hampered by the challenge of data availability \cite{strickland2022andrew, lee2022radio, whang2023data}.
Additionally, data providers, who are typically reluctant to share their data without incentives, may hinder progress and even lead to the termination of projects \cite{tenopir2015changes}.
Accordingly, as the adoption of the Internet of Things (IoT) broadens and data collection from edge devices intensifies, the discourse has shifted towards harnessing this data while protecting the personal information of data providers.
In this context, federated learning has emerged as a promising solution because it can offer improved artificial intelligence (AI) models in a way that data privacy is maintained despite utilizing valuable data from client devices \cite{mcmahan2017communication, kim2023collaborative}.
However, federated learning still faces challenges, including the lack of punitive measures for clients who deliberately disrupt the learning ecosystem \cite{bagdasaryan2020backdoor, gong2022backdoor}, potential issues associated with centralized learning such as the single point of failure problem \cite{ma2022federated}, and the inability for learners to claim and verify their ownership of locally generated models \cite{nguyen2021federated}.
Furthermore, it is particularly critical to address system heterogeneity in federated learning \cite{lee2022adaptive}, taking into account the diverse characteristics of the multitude of devices involved.

The integration of blockchain with federated learning is a rapidly evolving area of research, aimed at addressing the aforementioned limitations \cite{nguyen2021federated, qu2022blockchain, ali2021integration, zhu2023blockchain}.
However, there is a conspicuous scarcity of practical applications that have been rigorously tested within real-world contexts.
Notably, existing studies have predominantly focused on the theoretical facets of integrating these technologies, yet have not thoroughly examined the constraints and challenges associated with its practical deployment.
To bridge this research gap, in this paper, we develop a novel reference architecture for blockchain-enabled federated learning (BCFL).
This architecture is specifically designed to facilitate practical research and real-world implementation, thereby providing an actionable blueprint for future BCFL applications.
%Our approach addresses the aforementioned limitations by integrating blockchain technology with federated learning techniques.
To this end, we introduce an incentive system underpinned by smart contracts and employ decentralized identifiers (DIDs) for authentication.
Our main contributions include:

\begin{itemize}
\item Designing a BCFL reference architecture and implementing and verifying it in a practical Ethereum development environment,\footnote{Our decision to use Ethereum stems primarily from its extensive developer community and comprehensive library, which facilitates the most universal construction of decentralized applications (DApps) in the Web3 environment. Additionally, the widespread adoption of the Ethereum virtual machine (EVM) across many blockchain networks makes it easier to deploy smart contracts written in Solidity on various chains. This compatibility and ease of application greatly influenced our choice.}
\item Defining and conducting a comprehensive analysis of smart contract functions, stakeholder roles (e.g., job creators, evaluators, and trainers), and the use of interplanetary file system (IPFS) for sharing learning models among federated learning participants,
\item Proposing an extension architecture to support cross-device and cross-silo federated learning scenarios,
\item Developing a method for ID access and management for federated learning participants through integration with a DID access system, and 
\item Reviewing operating costs through a comparison of deployment costs in the Ethereum test network and the local simulation network of BCFL.
\end{itemize}

The rest of the paper is organized as follows.
Section \ref{sec:headings} provides an examination of the key terms and introduces some related works.
In Section \ref{sec:BCFL}, we present a detailed explanation of our proposed approach, including the overall workflow and the roles of each component.
Section \ref{sec:experiments} showcases the experimental results conducted in a real-world environment.
Finally, Section \ref{sec:conclusion} concludes the paper.

\section{Background and Related Work}
\label{sec:headings}
\subsection{Federated Learning}
The concept of federated learning was first introduced by Google in 2017 as a solution to the challenge of training machine learning models with distributed data \cite{mcmahan2017communication}.
Federated learning is a machine learning strategy in which multiple entities collaboratively train a shared model without needing to exchange their raw data.
More specifically, this approach processes data locally on individual devices, thereby eliminating the need for data collection or centralized storage in a server.
As a result, it can significantly reduce server resource usage and ensure data privacy \cite{kim2023collaborative, li2020federated, li2023survey, niknam2020federated}.
These benefits have contributed to the growing popularity of federated learning.
As its core, federated learning harnesses distributed computing power by enabling individual devices to contribute to the training of the shared model.
This approach not only maintains data privacy and security but also facilitates the development of models that are specifically tailored to the unique needs of each device.

Since its inception, federated learning has made substantial strides, with researchers proposing various techniques to tackle its challenges.
Early approaches relied on simple averaging algorithms to merge updates from multiple devices, but recent advancements have demonstrated that performance can be enhanced by the aid of client and data selection algorithms \cite{zhao2022federated, sattler2020robust}.
These developments have significantly improved the practicality and efficiency of federated learning, with numerous research results demonstrating its application in diverse fields such as healthcare, finance, and transportation.
Despite being in its early stages, federated learning holds immense potential as it offers a novel way to enhance machine learning model performance while preserving data privacy and security.
Given that federated learning involves local training on devices with different data distributions and quantities, it is essential to conduct research on addressing data heterogeneity.
Consequently, there is an active and vibrant research community focused on studying non-independent and identically distributed (non-IID) data environments \cite{zhao2022federated, sattler2020robust, zhu2021federated}.

Although federated learning is one of the most active research areas and holds great potential, practical deployment and commercialization are still in their early stages partly due to various technical challenges.
To fill the gap, as research expands and is applied to various fields, various open-source libraries are emerging to compare, analyze, and apply these new algorithms \cite{he2020fedml, beutel2022flower}.
It offers a new way to improve the performance of machine learning models while maintaining data privacy and security.
Federated learning is expected to emerge as an important technology in the coming years.

\subsection{Blockchain}
Blockchain technology is a distributed ledger technology that enables secure, transparent, and distributed transactions \cite{di2017blockchain}.
It has gained significant attention across various industries due to its innovative features and capabilities.
Blockchain creates a permanent and unalterable record of transactions, making it an ideal solution for industries requiring trust, security, and transparency.

Essentially, blockchain is a digital ledger of transactions that is maintained by a network of computers called nodes.
Each node in the blockchain network maintains a copy of the ledger, and all changes are verified by consensus among the nodes.
Once a transaction is recorded on the blockchain, it cannot be changed or deleted, thus ensuring data integrity and immutability \cite{politou2021blockchain}.

One of the most significant advantages of blockchain technology is its decentralized nature \cite{puthal2018blockchain}.
This eliminates the need for intermediaries or central authorities, allowing all transactions to be transparent and accessible to all network members.
Thanks to such an advantage, blockchain technology is being applied to various use cases requiring data integrity and traceability, such as financial transactions \cite{nofer2017blockchain, albayati2020accepting}, supply chain management \cite{saberi2019blockchain, cole2019blockchain}, voting systems \cite{huang2021application}, identity management \cite{dunphy2018first, liu2020blockchain}, and healthcare \cite{zhang2018blockchain}.

In conclusion, blockchain technology is an innovative and disruptive technology that provides safe and distributed solutions to various industries.
It is an area of active research and application due to its unique features that make it an ideal solution for industries that require trust, security, and transparency.
It is also noted that the intersection between AI and blockchain technology has the potential to transform various industries by enhancing their security, transparency, and overall efficiency.

\subsection{Blockchain-Enabled Federated Learning (BCFL)}
% esgoh : Reviewer 2, Concern 3
BCFL, an emerging paradigm in machine learning, has attracted interest due to its potential in various areas, such as IoT and healthcare applications \cite{wang2021blockchain, li2020blockchain}.
By integrating the principles of federated learning and the security features of blockchain technology, BCFL facilitates data to be collected and processed locally on individual devices rather than being stored centrally, with the secure and transparent transaction recording of blockchain technology.
Thereby, this combination allows for secure and efficient data and model sharing between multiple parties without a central authority, potentially overcoming challenges associated with traditional machine learning methods, such as data privacy and security issues.

Specifically, within this framework, the role of blockchain lies in decentralizing the traditional federated learning structure, thereby eliminating the single point of failure issue associated with a central server and ensuring data immutability. This advancement yields technical benefits that enhance trustworthiness and transparency among participants. As discussed in \cite{nguyen2021federated}, the integration of blockchain with federated learning could revolutionize the conventional federated learning structure, by transforming it into a decentralized federated learning ecosystem that can safeguard personal information without the reliance on a central server. In \cite{myrzashova2023blockchain}, it is also pointed out that the susceptibility to errors in the aggregation of global models by a centralized federated learning server is a concern, suggesting that adopting BCFL could be one alternative solution.
%Additionally, challenges in the accuracy of aggregating global models by a centralized federated learning server are noted in \cite{myrzashova2023blockchain}.
Consequently, by employing BCFL, it is possible to address these concerns, offering an enhanced degree of security and private data protection in a decentralized fashion.
%As an example, this can be particularly beneficial for smart healthcare applications, where ensuring the confidentiality and integrity of sensitive data is paramount.

A notable example of BCFL is TrustFed \cite{ur2021trustfed}, a cross-device scenario-based framework designed to provide fairness and trust to participants.
It is implemented using blockchain smart contracts and statistical anomaly detection techniques.
Not only this, there are various other studies that emphasize the importance of data preservation \cite{jia2022blockchain, qi2021blockchain, singh2022framework, zhu2022blockchain, qu2020decentralized}.
Moreover, the synergy between federated learning and blockchain finds significant applications in various network environments.
For instance, the authors in \cite{rahmadika2021unlinkable} explored cross-silo federated learning to ensure privacy protection using cryptocurrency in edge networks.
In a similar vein, the study in \cite{nguyen2021federated} introduced a fundamental concept of a system that combines conventional federated learning with blockchain, thereby proposing a fresh paradigm with potential application areas in mobile edge computing (MEC) networks.

% esgoh : Reviewer 1, Concern 5
Beyond the aforementioned applications, BCFL also manifests significant potential within medical services in distributed environments. The work presented in \cite{egala2023fortified} aims to enhance data availability, security, and transparency by integrating a distributed data storage system (DDSS), blockchain, and hybrid computing. It is noteworthy that this methodology notably diverges from our BCFL reference architecture, which will be detailed later, wherein we emphasize a modular approach, structuring the specific roles that stakeholders should undertake in accordance with the systematic workflow intrinsic to BCFL. This design highlights the scalability and flexibility of our reference architecture, rooted in its modular fashion.

Furthermore, employing BCFL introduces another significant advantage: the implementation of automated reward mechanisms via smart contracts. This approach can help deter potential malicious participants commonly encountered in traditional federated learning ecosystems, including those who contribute counterfeit data during the training process. The authors in \cite{bao2019flchain} highlight a critical gap in traditional federated learning systems, namely the absence of adequate incentives to encourage the sharing of decentralized training data and computational resources. To address this and establish a decentralized, publicly auditable federated learning ecosystem founded on trust and incentives, they recommend adopting blockchain technology.
Similarly, the authors in \cite{gao2022fgfl} underscore the necessity of reasonable incentives, noting that without them, participants may hesitate to engage in the learning process. Moreover, the study emphasizes the pressing demand for incentive strategies to deter malicious participants aiming to degrade the model's performance. In response to this challenge, they suggest an incentive scheme that assesses participants based on reputation and contribution metrics.

Building on the concept of incentives in BCFL, recent developments have put a spotlight on the intricacies of the incentive and reward mechanism.
One of the complexities arises from the distributed nature of BCFL, where participants contribute to learning for their associated edge devices, making it very difficult to directly monitor the behavior of participants.
To address this issue, the authors in \cite{toyoda2020blockchain} leveraged competition theory from the field of economics to provide a mathematical and systematic solution to the reward mechanism.
This innovative solution exemplifies the ongoing efforts to refine BCFL's incentive structures.
Simultaneously, the urgency for more secure and efficient data-sharing methods in a variety of industries, including healthcare, finance, and e-commerce, propels the advancement of BCFL technologies.
The burgeoning data volume and escalating privacy concerns make this decentralized approach an increasingly critical solution.
Anticipated future developments in BCFL are likely to include the introduction of new algorithms and protocols aimed at enhancing security and efficiency further.
As more organizations recognize the potential benefits of this approach, BCFL is poised to gain mainstream acceptance in the realms of machine learning and data sharing.
Nonetheless, the path towards widespread BCFL implementation is not without hurdles, as evidenced in that available resources to realize BCFL remain relatively limited 
despite the significant advances in the architecture design \cite{lo2023toward} and open-source projects \cite{cai20202cp, dias2022impact}.

\subsection{Decentralized Identifiers (DID)}
DID, gaining attention in recent years, is a method for managing and protecting digital identity in a decentralized manner \cite{avellaneda2019decentralized, bai2022decentralized}.
It employs a unique identifier to create a verifiable, reliable, and tamper-proof digital identity, which is independent of control by any central authority.

One crucial feature of DID is the distribution of ownership and control of digital identities across multiple parties, counteracting the control typically held by a single organization or legal entity.
This distribution is possible thanks to blockchain technology, providing a distributed, transparent mechanism for managing identities and transactions.
Within this framework, privacy stands out as one of DID's most significant aspects.
Specifically, unlike traditional identity management systems where central authorities collect and store personal information, potentially leading to data breaches and privacy violations, DID ensures privacy by enabling individuals to manage their personal data and decide when and with whom to share.
As a result, the data can be shared only with trusted parties when necessary.

In DID systems, personal information is stored in a distributed, encrypted format, presenting a higher level of privacy and security compared to conventional identity management systems.
As the digital landscape evolves, there is a growing demand for secure, decentralized identity management solutions.
DID is emerging as a promising solution, allowing individuals greater control over their personal information, thereby enhancing their privacy and security.

% esgoh : Reviewer 2, Concern 2
Moreover, the integration of DID with verifiable credentials (VCs) offers enhanced privacy and security compared to centralized systems, granting users more control and preventing large-scale data breaches common in centralized databases. These features align with the core objectives and values of BCFL, which emphasizes learning within distributed environments and reinforcing individual data ownership.

Notably, regarding security threats, DID with VCs can be instrumental in mitigating threats such as Sybil attacks, where malicious entities create and distribute some fake identities to compromise systems. Specifically, DID coupled with VC stands as a forward-looking identity authentication method suited for distributed environments, synergizing with BCFL and similar cutting-edge, advanced technologies.

\subsection{Interplanetary File System (IPFS)}
IPFS is a distributed peer-to-peer (P2P) file system that addresses limitations of the traditional centralized internet system, thereby providing a technical solution for secure and rapid transmission of distributed data.
IPFS usually employs a hash-based file system (HFS) for file storage and connects with the blockchain technology to ensure file uniqueness and enhance the security of distributed data storage \cite{nyaletey2019blockipfs, naz2019secure}.

IPFS can be utilized for a variety of purposes, the most common being file sharing and distributed web hosting.
The reason for this is that IPFS allows files to be shared using the distributed technology of the blockchain without needing to locate the original files.
Additionally, IPFS can also host websites in a distributed manner similar to a content delivery network (CDN), utilizing its distributed technology.
These innovative solutions hold significant potential to enhance the security and safety of the internet.

\section{BCFL Reference Architecture}
\label{sec:BCFL}
In this section, we first outline the structure of our proposed architecture, which comprises two phases encompassing six stages, and provide detailed information about each stage.
We then engage in an in-depth discussion on the smart contract functionalities within the architecture.
Following this, we explore the cross-device and cross-silo scenarios and discuss the functions of stakeholders along with the design of other modules.
Lastly, we conclude this section with a discussion on the concept of the DID and VC certification system.

\subsection{Overview of the Six-Stage BCFL Workflow}

\begin{figure}[!t]
\centering
\includegraphics[width=3in]{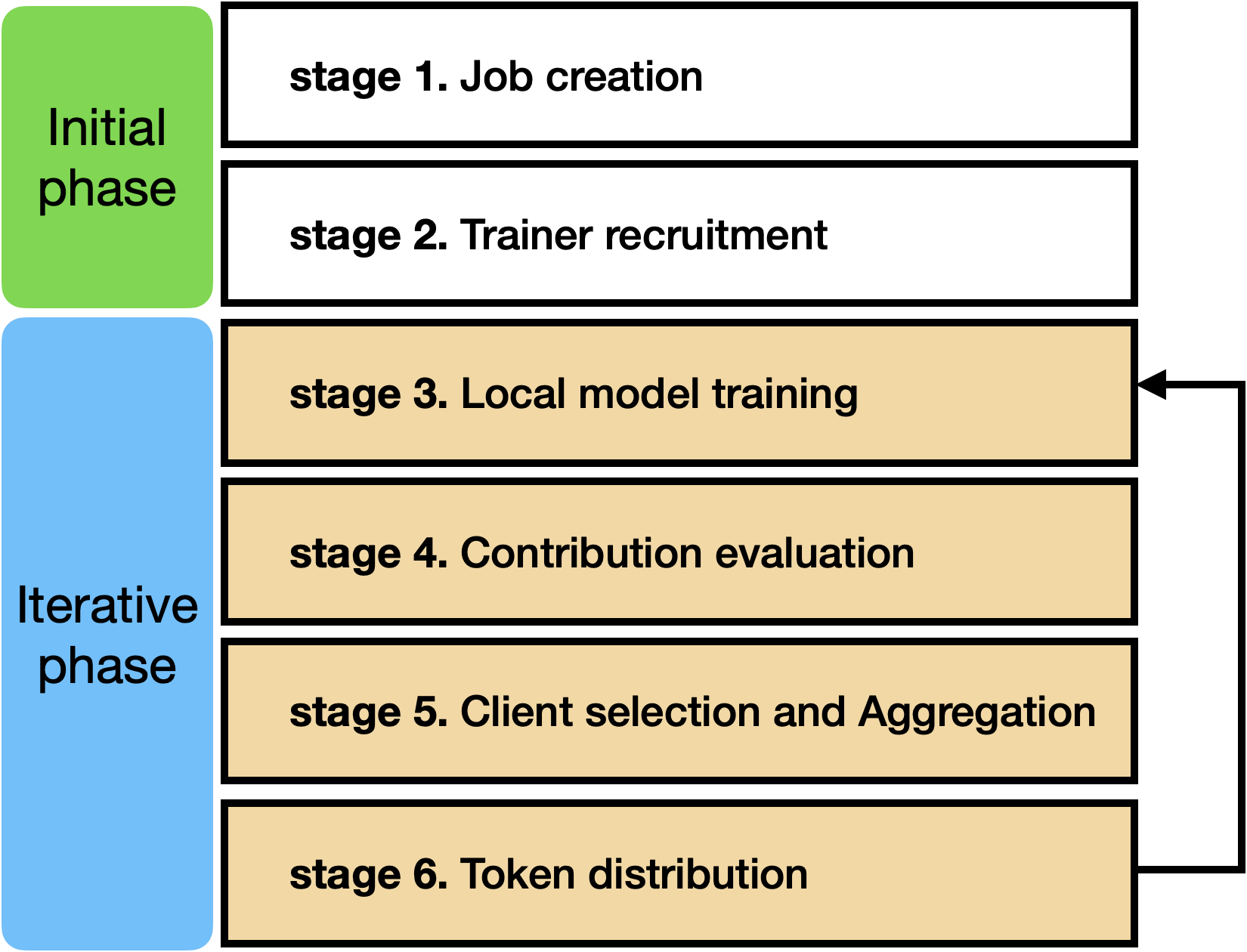}
\caption{Overall workflow of BCFL, segmented into two phases encompassing six stages.}
\label{fig:fig1}
\end{figure}

\begin{figure*}[!t]
\centering
\includegraphics[width=\linewidth]{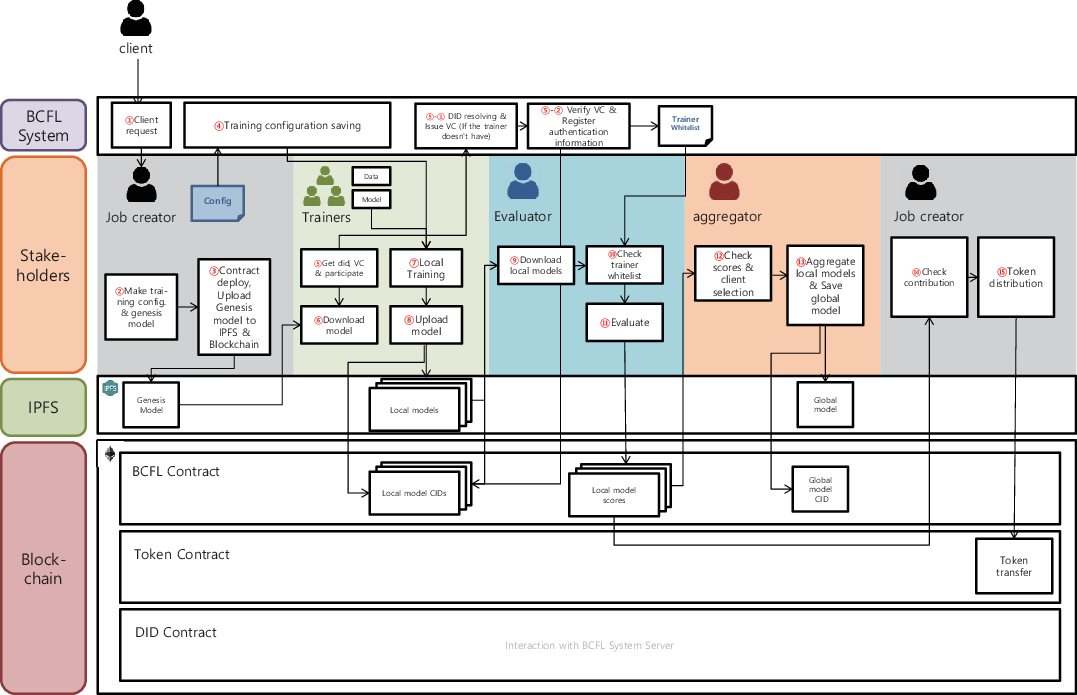}
\caption{Detailed representation of the complete BCFL system architecture.}
\label{fig:fig2}
\end{figure*}

Fig. \ref{fig:fig1} offers an at-a-glance view of the six stages encompassed within our BCFL workflow, which is categorized into two distinct phases.
The initial phase includes job creation and trainer recruitment, while the second phase involves the iterative execution of four stages: training, evaluation, client selection and aggregation, and token distribution.
These latter four stages are repeated for a predetermined number of global rounds, ensuring comprehensive training and evaluation.
Transitioning to a detailed perspective, Fig. \ref{fig:fig2} presents a comprehensive diagram that cross-references the key entities with their respective roles. 
Horizontally, it categorizes the essential entities into four distinct types: the BCFL system, stakeholders, IPFS, and blockchain.
Vertically, it identifies the types of participants: job creators, trainers, evaluators, and aggregators.
This intricate portrayal not only underscores the interactions between different entity types but also illuminates the layered complexity of roles within the proposed BCFL ecosystem.

We now turn our attention to the essential entity types themselves, and thereafter, we will discuss the details of each of the six stages.
Note that these stages represent the functional steps through which our proposed BCFL reference architecture operates, delineating the sequence of processes that the job undergoes from creation to completion, inclusive of the dissemination of rewards.

\begin{itemize}
\item The first entity type, the BCFL system, provides users with fundamental functionalities via web applications or similar platforms.
It includes providing an environment for trainer recruitment through the user interface, supporting the registration process, and more.
\item The second entity type comprises stakeholders, who are the actual users that participate in the learning process.
Within our architecture, these participants include job creators, trainers, evaluators, and aggregators.
Their specific roles and functions at each stage are depicted in Fig. \ref{fig:fig2}.
\item The third entity type, IPFS, serves as the repository for storing the outcomes of the learning process.
\item Lastly, the blockchain constitutes the fourth entity type, serving as the foundation for smart contracts that oversee the learning process.
Included within this are the BCFL contract, which orchestrates federated learning-related functions; the Token contract, which manages transactional functions based on ERC-20 tokens; and the DID contract, which handles DID authentication functions.
\end{itemize}
Having outlined the essential entity types, we will proceed in the following subsections to detail the six stages of the BCFL workflow, referencing the pertinent aspects of Fig. \ref{fig:fig2}.

\subsubsection{Stage 1: Job creation}
% esgoh senction B Job creation 단계는 \ref{fig:fig2}에서 1단계부터 4-1단계에 해당합니다.
A client initiates a BCFL task, which prompts the job creator to generate a quote based on the client's specifications.
This quote details the type of deep learning model to be used, the configuration of learning hyperparameters, the desired number of trainers, the number of global rounds, the genesis model,\footnote{Inspired by the term `genesis block,' which denotes the first block mined in any cryptocurrency, we have coined the term `genesis model' to refer to the initial model to be distributed under the BCFL framework.} etc.
The job creator then deploys the genesis model and registers the task.
%esgoh(1) : 배포 과정이 밑에 있어서 삭제 to the BCFL system and registers the task.
The deployment process entails uploading the model's parameters to IPFS and then recording the returned content identifier (CID) on the blockchain.
An essential part of this process is interacting with the smart contract through a cryptographic wallet, which is necessary for recording the details on the blockchain.
%esgoh 이 배포 과정은 IPFS에 먼저 모델의 파라미터를 업로드 한 뒤, 반환된 content id(cid)를 블록체인에 기록하는 순서로 이루어집니다. 이것을 블록체인 상에 기록하기 위해서는, 암호화 지갑을 이용하여 smart contract와의 상호작용이 필수적으로 필요합니다.
It should be noted that the client who requests the job might be a distinct entity or might also take on the role of job creator.
For visual reference, this job creation stage is depicted in Fig. \ref{fig:fig2}, ranging from block 1 to block 4.

\subsubsection{Stage 2: Trainer recruitment}
% esgoh senction B 이 과정은 \ref{fig:fig2}에서 5-1부터 5-1-2까지의 단계에 해당합니다.
Trainers should be able to review the training task via a dedicated web application and estimate the potential benefits (tokens) they can earn.
Specifically, based on the training configuration created by the job creator (block 4), trainers assess their suitability for the task and, if appropriate, request participation through the application.
The BCFL system manages the list of applicants and continually monitors the learning process outcomes.
The system also filters out malicious trainers by scoring their behavior and creates a whitelist of approved trainers, thereby maintaining the integrity of the learning process.
This persistent monitoring and filtering are crucial to safeguard against the recurrent negative influence of malicious trainers, preventing their participation in other federated learning tasks.
For identity verification, the trainers can use a VC JSON web token (JWT) or, alternatively, issue a new DID and VC. 
During this process, the BCFL system validates the participants' information and securely stores the authentication data, enabling trainers to accrue corresponding bonus points.
% esgoh senction B  Trainer들은 job creator에 의해 작성된 training configuration(4-1단계)을 참고하여 본인이 해당 task에 적합한지 판단한 후 web application을 통해 참여를 요청합니다. 이 때 학습에 지원한 사람들의 list는 BCFL system이 관리합니다. BCFL system은 학습의 결과를 지속적으로 모니터링하며, 악의적인 행위를 하는 trainer를 score를 통해 필터링하고, 악의적인 trainer를 제외한 나머지 trainer의 list를 작성해서 whitelist로써 관리합니다. 이 과정은 꾸준히 악영향을 끼치는 trainer를 선별하여 다른 FL task에 해당 trainer가 참여하지 못하도록 방어하기 위해 필요합니다.
This trainer recruitment stage, including the verification and whitelisting process, is depicted in Fig. \ref{fig:fig2}, ranging from block 5 to block 5-2.

\subsubsection{Stage 3: Training}
% esgoh senction B 이 과정은 \ref{fig:fig2}의 5-2부터 7까지에 해당합니다.
Once the trainer recruitment stage is complete and all trainers are ready, the smart contract updates the training status to the training phase and signals the start of training.
The trainers then begin by downloading the genesis model.
This step is achieved by first invoking a function in the smart contract that returns the genesis model's CID, followed by the acquisition of the model parameters from IPFS using that CID.
With the genesis model in hand, the trainers commence training with their own datasets.
% esgoh senction B  Genesis model을 다운로드 하는 방식은, 먼저 smart contract에서 genesis model cid를 반환하는 함수를 call한 뒤 IPFS에 해당 cid를 기입하여 모델 파라미터를 다운로드합니다. 
Following this, they register their local model updates on both the IPFS and the smart contract.
The registration process mirrors the job creator's initial upload of the genesis model, as depicted in block 3 in Fig. \ref{fig:fig2}, with the addition of key supplementary information logged in the smart contract.
This information serves to identify which trainer contributed to the update and the specific global round it pertains to, among other details.
This training stage spans from block 6 to block 8 in Fig. \ref{fig:fig2}.
% esgoh senction B  7단계에서 trainer들이 학습한 local model을 업로드하는 방식은 3단계에서 job creator가 모델을 업로드하는 방식과 동일하지만, smart contract에 부가적으로 기록하는 정보가 있습니다. 이 정보에는 어떤 트레이너가 학습한 모델인지, 몇 번째 global round에 해당하는 모델인지 등이 해당합니다.

\subsubsection{Stage 4: Evaluation}
% esgoh senction B 이 과정은 \ref{fig:fig2}의 8단계부터 10단계 까지에 해당합니다.
Upon completion of the training stage in a global round, the evaluator reviews the status of the smart contract to decide if the evaluation stage can commence.
At the beginning of this stage, the evaluator downloads the local model parameters submitted by the trainers.
This process exactly mirrors how local trainers acquire the genesis (or global) model, as depicted in block 6.
The evaluator then carries out the evaluations with pre-prepared data.
% esgoh senction B Evalution을 진행하기 위해, evaluator는 가장 먼저 trainer들이 제출한 local model parameter들을 다운로드 받습니다. 이 과정은 5-2 단계에서 local trainer들이 genesis model혹은 global model을 얻는 방식과 완전히 동일합니다. 
Before this, the evaluator retrieves DID authentication client information from the BCFL system and awards bonus points to trainers verified through DID (who can be called DID-authenticated trainers).
For awarding these bonus (contribution) points, a variety of algorithms are available, and an appropriate method can be chosen for each specific BCFL task.
When it comes to recording the trainers' scores, the evaluator compiles the information, including the score, the trainer's wallet address, and the CID, and then registers it in the smart contract.
This structured approach to compiling and recording ensures that both the BCFL system and the aggregator can easily reference the scores.
After the evaluations are finalized, the evaluator documents the model-specific scores for each trainer in the contract.
This evaluation stage extends from block 9 to block 11 in Fig. \ref{fig:fig2}, covering the process from the downloading of model parameters to the documentation of the evaluation outcomes.
% esgoh senction B Trainer의 점수를 기록할 때에는, 점수와 trainer의 지갑 주소, cid를 매칭 시켜서 구조체 형태로 등록함으로써 BCFL system 및 aggregatior가 이것을 참고할 수 있도록 합니다. 

\subsubsection{Stage 5: Client selection and aggregation}
This stage corresponds to blocks 12 through 14 in Fig. \ref{fig:fig2}.
It is important to note that the client selection process, which pertain to blocks 11, is optional part and are not strictly mandatory for implementing the our BCFL architecture.
If the client selection process is undertaken, the next round's participants can be determined based on scores recorded in the smart contract, according to a specific pre-determined or predefined protocol that may vary in definition.
Subsequently, 
% esgoh senction B 이 단계는 \ref{fig:fig2}의 11단계부터 13단계에 해당합니다.
% esgoh senction B 11단계와 12단계는 optional하며, 필수적이지 않습니다. 이 두 단계가 client selection에 해당합니다. client selection을 진행하는 경우에는 smart contract를 통해 기록된 score를 참고하여 다음 라운드부터 탈락시킬 trainer를 결정합니다. 
% esgoh senction B 13단계에서는 aggregator가 본격적으로 trainer들이 학습한 local 모델을 aggregation하는 작업을 시작합니다.
the aggregator interacts with the smart contract to retrieve lists of the trainers, model CIDs, and contribution points for the recorded models.
Employing algorithms like FedAvg \cite{mcmahan2017communication}, the aggregator then synthesizes the local models provided from the trainers into a unified global model.
% esgoh senction B The resulting global model is documented in both IPFS and the smart contract. 이 과정 역시 7단계 및 3단계에서의 모델 업로드 방식과 동일합니다. 
This global model is then recorded in both IPFS and the smart contract, a process that can be implemented in the same manner as model uploads in blocks 3 and 6.
Additionally, at the end of the aggregation stage within a round, the aggregator assesses clients based on their scores and records the list of trainers who qualify for the next round in the contract, thereby notifying them of their continued participation.

\subsubsection{Stage 6: Token distribution}
% esgoh senction B 이 단계는 \ref{fig:fig2}의 14, 15단계에 해당합니다.
In this stage, the job creator utilizes a pre-deployed token contract to distribute tokens to trainers, with the distribution amounts determined by the score list.
These tokens are intended to act as stakes in the final global model.
These steps are represented by blocks 14 and 15 in Fig. \ref{fig:fig2}.

\subsection{Smart contract}

The smart contract oversees the overall workflow and handles critical information throughout the process.
Table \ref{tab:table1} enumerates the functions required in the contract for the BCFL cross-device scenario.\footnote{In this study, we will consider not only a cross-device scenario but also a cross-silo scenario. Detailed explanations of these scenarios will be provided later in Section \ref{sec:scenarios}.}
Below is a detailed description of each function.

\begin{table*}[!t]
\caption{Function classification and description for the BCFL cross-device scenario in the smart contract.}
\setlength{\tabcolsep}{3pt}
\centering
\begin{tabular}{lll}
\toprule
	Function name & Key actions & Detailed description\\
\midrule
	\multirow{3}{*}{Round control}
	& Current round & Return current round\\
	& Round remaining seconds & Return remaining seconds of entered round\\
	& Round skipping & Set round to next\\
\midrule
	\multirow{4}{*}{Client selection}
	& Get current trainers & Return current round's trainer list\\
	& Check valid trainer & Returns the status of whether the trainer can participate in learning in the input round\\
	& Set current trainers & Register a list of trainers to participate in the current round\\
	& Change number of updates & Change the number of updates (number of trainers) that must be registered for that round\\
\midrule
	\multirow{4}{*}{Evaluation} & Score saving & Save local modes' contribution score\\
	& Get model score & Return local model's contribution score\\
	& Set evaluation completion flag & Register a specific round as evaluated\\
	& Set evaluator & Change evaluator as entered account\\
\midrule
	Global model saving & Save global model & Save global model's CID\\
\midrule
	\multirow{2}{*}{Training} & Get global model & Return entered round's global model\\
	& Add local model update & Save the local model CID for a specific round\\
\midrule
	\multirow{4}{*}{Token distribution} & Count tokens per round & Returns the total amount of tokens distributed in the input round\\
	& Count total tokens & Return the total amount of distributed tokens\\
	& Set tokens per round & Record the amount of tokens used in the entered round\\
	& Transfer tokens to address & Send tokens to a specific wallet address\\
\midrule
	\multirow{5}{*}{FL progress management} & Training phase & Returns the current training phase (training, evaluation, aggregation, pending, etc.)\\
	& Get number of updates & Returns the total amount of updates for the current round\\
	& Wait trainers & Waiting for a trainer whose learning has not finished\\
	& Check trainer's update & Returns the status after checking if the received trainer has finished updating in the round\\
	& Updates in round & Returns the CID list of local models of the received round\\
\bottomrule
\end{tabular}
\label{tab:table1}
\end{table*}

\subsubsection{Round control}
In BCFL, there generally is not a separate server to manage the learning process.
Hence, the smart contract needs to handle the registration and management of information in a round and its duration.
The round control module contains these functions, enabling stakeholders to continuously monitor the current round and the remaining time during the learning process.
A round is deemed completed once the evaluation and aggregation for that round are finished, and the process then advances to the next round.

\subsubsection{Client selection}
In client selection, managing the trainers for each round based on evaluation results is essential.
After the aggregator performs client selection in a specific round, the system needs to register the wallet addresses of eligible trainers.
This function should be set up such that the trainers can easily verify it.
At the start of the round, trainers can check their status to see if they are valid participants, and based on the status value they receive, they can decide whether to continue participating in the BCFL process.

\subsubsection{Evaluation}
The evaluator assesses the performance of the local models submitted by trainers and records the evaluation results in the smart contract.
The smart contract should be capable of storing the CID of the local model, its corresponding score, and the trainer information for the model.
To ensure access control, the evaluator's access to the evaluation score should be restricted using measures like solidity modifiers.
Once the evaluation of all local updates in a specific round is completed, the evaluation completion status value should be changed to allow other trainers to progress to the next training stage.
While it is possible to change the evaluator, such changes are restricted to unavoidable cases.

\subsubsection{Global model saving}
After the client selection process is completed, the aggregator should be able to upload the CID of the aggregated global model through the contract.

\subsubsection{Training}
During the training process, trainers download the global model, train it using their own data, and subsequently upload the trained model to IPFS.
Consequently, the smart contract should include functions to store and retrieve the CID of the global model and the updated CID of the local model for the corresponding round.

\subsubsection{Training initiation}
The job creator's role includes defining which model to use as the genesis model at the beginning of the training process, and recording the CID of the initial version of the model, which is uploaded through IPFS, in the contract.
This ensures that all participants have access to this model for training.

\subsubsection{Token distribution}
Trainers seek to be rewarded for their contributions.
The token distribution function facilitates this process by linking ERC-20 and other token interface contracts, allowing the issuance of tokens that can be traded on an actual network.
These tokens hold value, such as operating as a stake in the final global model.
These tokens hold value, serving various purposes such as acting as a stake in the final global model.

\subsubsection{FL progress management}
In addition to the functions listed above, the following functions are crucial for effectively managing the progress and status of the federated learning training:

\begin{itemize}
\item After all the trainers are recruited, the smart contract receives a notification signaling that the federated learning training is ready to start.
\item A status check is performed to verify whether a particular trainer has successfully uploaded the CID of its local model for the current round. This check ensures that there are no abnormalities in the training process and that each trainer's contribution is accounted for.
\item To evaluate the contribution of the local models for the corresponding round, a list of model CIDs that have completed training is necessary. This list helps in assessing the impact of the trained models and further analysis and aggregation.
\end{itemize}

\subsection{Cross-device and cross-silo scenarios}
\label{sec:scenarios}

\begin{figure*}[!t]
\centering
\includegraphics[width=6in]{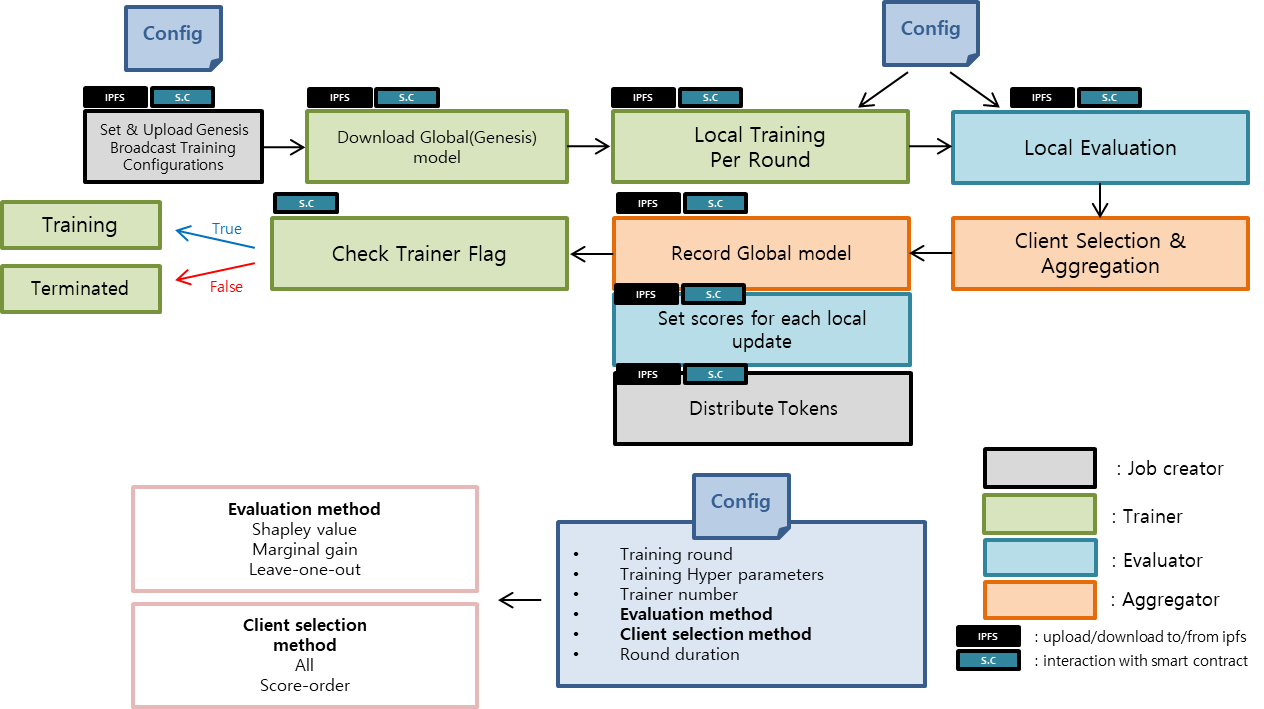}
\caption{Workflow illustration of the cross-device scenario in BCFL.}
\label{fig:fig3}
\end{figure*}

Fig. \ref{fig:fig3} depicts the overall workflow of the cross-device scenario in BCFL.
The roles and functionalities of each stakeholder are represented in different colors, and badges are attached to indicate interactions with IPFS and smart contracts.

The job creator initiates the BCFL process by reviewing the received tasks and establishing the training configuration.
This includes the total number of global rounds, training hyperparameters, the number of trainers, evaluation methods, client selection methods, and round duration.
Once participant recruitment is completed by the job creator, the genesis model is uploaded to IPFS.
The job creator then registers the CID of the uploaded genesis model in the smart contract, providing access to all participants.

\begin{figure*}[!t]
\centering
\includegraphics[width=6in]{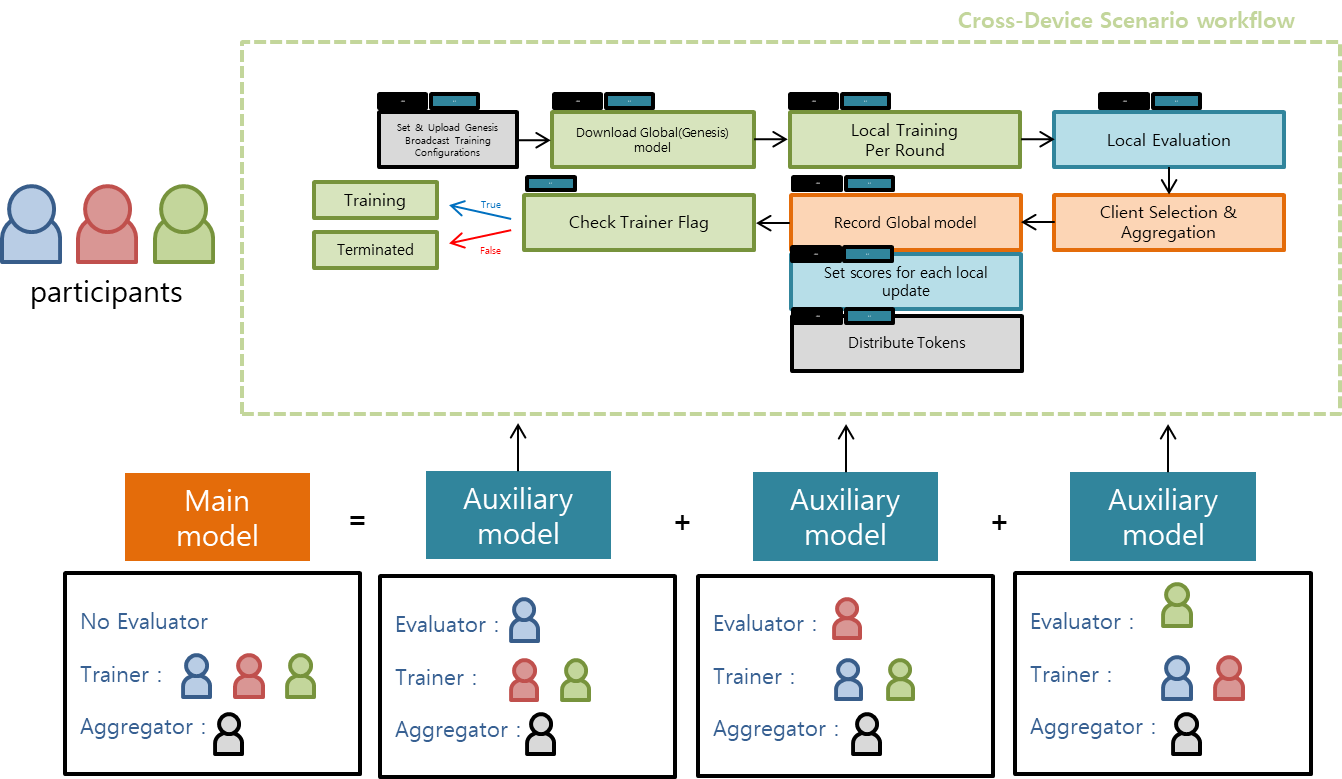}
\caption{Workflow illustration of the cross-silo scenario in BCFL.}
\label{fig:fig4}
\end{figure*}

Fig. \ref{fig:fig4} illustrates the workflow of the cross-silo scenario, which adapts and enhances the model presented in \cite{cai20202cp}.
This scenario presupposes the participation of three entities, with the final global model emerging as an aggregation of their respective auxiliary models.
Each participant cyclically takes the evaluator role within its auxiliary model while the others serve as trainers.
As there is no distinct evaluator for the main model, the cumulative evaluation results from each auxiliary model are used as weights.

The aggregator is responsible for aggregating the auxiliary models and executing the final aggregation for the main model.
As illustrated in Fig. \ref{fig:fig4}, a third-party entity, not considered as a participant, is chosen as an example.
To identify an eligible client for aggregation, the BCFL system, which manages the whitelist and client requests, should maintain a whitelist of authenticated users, such as those with DIDs, thus selecting users who consistently contribute to the training.
If an aggregator is chosen from among the participants, strategic variations can be introduced, which can be done by randomly selecting an aggregator from the clients and assigning the role to the client with a high contribution during the training process.

\subsection{Stakeholders' functions and other module designs}
In this work, we categorize stakeholders into two distinct groups based on the scenario.
In the cross-device scenario, we identify four types of stakeholders: job creator, evaluator, trainer, and aggregator.
Alternatively, in the cross-silo scenario, we identify three types of stakeholders: job creator, participant, and aggregator.
As described later, we propose that each participant in the cross-silo scenario can perform both the roles of an evaluator and a trainer.

\subsubsection{Job creator}

\begin{figure}[!t]
\centering
\includegraphics[width=3in]{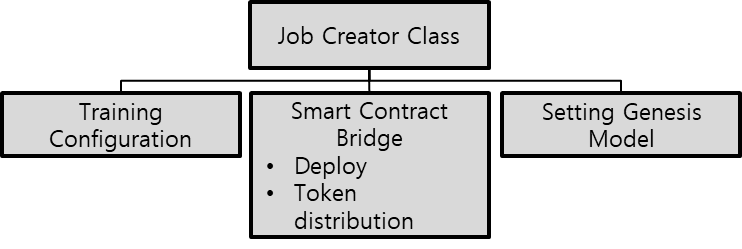}
\caption{Structure of the job creator class module within the context of the cross-device scenario.}
\label{fig:fig5}
\end{figure}

The job creator can either be the entity commissioning the task or a client who receives requests from external organizations and prepares estimates.
The job creator is primarily responsible for the creation part of the BCFL task.
These responsibilities include creating the training configuration, utilizing the BCFL cross-device scenario smart contract along with network standard interfaces, such as ERC-20 token contracts, deploying DID contracts, and setting up the genesis model.
As outlined in Fig. \ref{fig:fig5}, the key responsibilities of the job creator are as follows:
\begin{itemize}
	\item \textbf{Training configuration}: This encompasses details such as the number of global rounds, the number of trainers, and the evaluation and client selection algorithms. This information is disseminated to the stakeholders through either the BCFL system or the smart contracts.
	\item \textbf{Smart contracts}: The job creator deploys the requisite smart contracts essential for training. This encompasses the BCFL contract, the token contract using network standard interfaces, and the DID contract. The job creator accesses the contract after each global round's completion to verify the status and distribute tokens to the trainers.
	\item \textbf{Setting genesis model}: The job creator designs the deep learning network in accordance with the desired objectives and uploads the model to IPFS. This information, once documented in the smart contract, becomes accessible to all stakeholders.
\end{itemize}

\subsubsection{Evaluator}

\begin{figure}[!t]
\centering
\includegraphics[width=\linewidth]{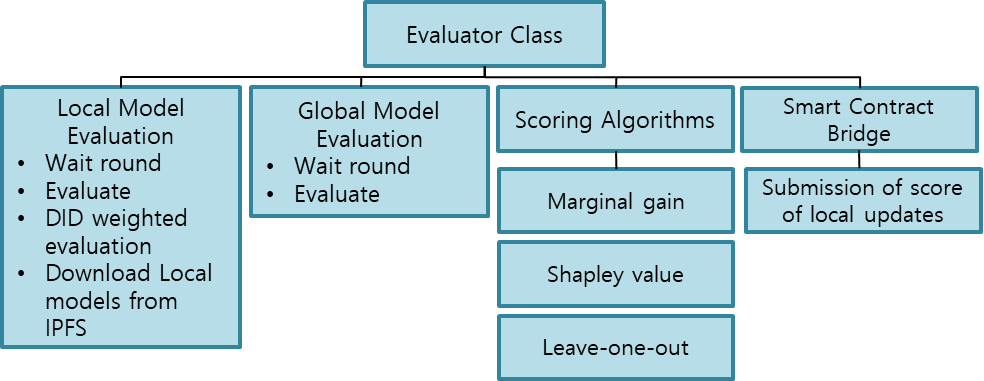}
\caption{Evaluator class module structure in cross-device scenario.}
\label{fig:fig6}
\end{figure}

Presumed to be a node equipped with a suitable test data set, the evaluator is responsible for evaluating local models.
It is designed to be capable of receiving rewards for providing evaluation data and contributing to the training process.
As outlined in Fig. \ref{fig:fig6}, the evaluator class consists of modules for `local model evaluation,' `global model evaluation,' `scoring algorithms,' and `smart contract bridge.'
These modules fulfill the following roles:
\begin{itemize}
	\item \textbf{Local model evaluation}: Once all the trainers have completed their local training, the evaluation phase begins. At this stage, the evaluator identifies the trainers who are authenticated via their DIDs and awards additional scores to them for reward purposes. Additionally, to perform the evaluation, the evaluator needs to access the smart contract to retrieve the CIDs of the models that the trainers have completed.
	\item \textbf{Global model evaluation}: To gauge the performance of the final model, an evaluation of the global model is required. This evaluation can serve as a comparative measure, indicating the extent to which the local models, trained by the trainers, contribute to the global model.
	\item \textbf{Scoring algorithms}: There are various types of algorithms for evaluation, with Shapley value and leave-one-out being commonly mentioned in the field of federated learning. The contribution in federated learning can be measured by the marginal value for a trainer's contribution to the global model, or by contribution estimation algorithms such as Shapley value or leave-one-out, in accordance with the guidelines specified in the training configuration created by the job creator. The chosen algorithm should align with the training's objectives and characteristics, and also allow for straightforward module design of tailored algorithms specific to the training task.
	\item \textbf{Smart contract bridge}: The submission of the completed local update CIDs to the blockchain is facilitated through the smart contract bridge. This bridge serves as a connection between the evaluation process and the blockchain, ensuring that all evaluation results are accurately recorded and readily accessible for subsequent stages of the BCFL process.
\end{itemize}

\subsubsection{Trainer}

\begin{figure}[!t]
\centering
\includegraphics[width=3in]{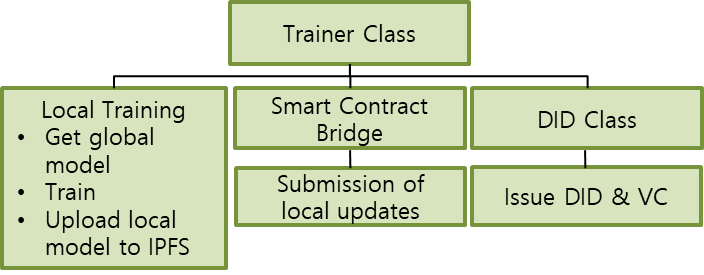}
\caption{Trainer class module structure in cross-device scenario.}
\label{fig:fig7}
\end{figure}

Trainers conduct the training in accordance with the provided training configuration and record the results in IPFS and smart contracts.
They also have the discretion to decide whether or not to use their personal information to obtain DID and VC.
Fig. \ref{fig:fig7} outlines the structure of the trainer class module, each detailed as follows:
\begin{itemize}
	\item \textbf{Local training} During the local training phase, trainers need to download the global model before they start their training.
The access CID for the global model is specified in the smart contract, and trainers access it to download the model from IPFS.
The training is conducted based on the training configuration provided by the job creator.
Upon completing their training, each trainer uploads their updated local model to IPFS, records it on the blockchain via the contract, and then waits for the evaluation by the evaluator to be completed.
After the evaluation stage, trainers can check whether they have been eliminated by the client selection algorithm at the start of the new round.
After verifying their status, they proceed to the next round.
    \item \textbf{Smart contract bridge}: Trainers are responsible for uploading the CIDs of their locally updated models to the blockchain through the smart contract. This bridge facilitates the recording of updates and contributions made by each trainer to the global model.
    \item \textbf{DID class}: Trainers have the discretion to obtain DID and VC voluntarily. If trainers choose to obtain DID and VC, they interact with the BCFL system, which includes authentication logic. The DID class manages these requests to the BCFL system, and the BCFL system, in turn, maintains a whitelist of authenticated trainers and manages it for the corresponding BCFL task.
\end{itemize}

\subsubsection{Aggregator}

\begin{figure}[!t]
\centering
\includegraphics[width=3in]{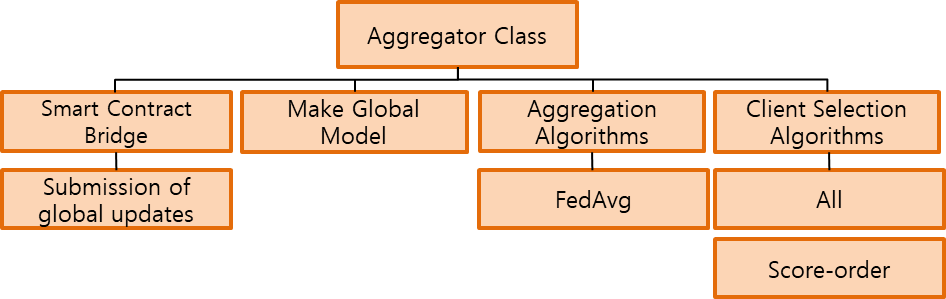}
\caption{Aggregator class module structure in cross-device scenario.}
\label{fig:fig8}
\end{figure}

The aggregator performs global model aggregation and receives additional rewards accordingly.
The criteria for selecting the aggregator can rely on the whitelist furnished by the BCFL system, which allows for the selection of clients beyond the pool of training participants.
Alternatively, the flexibility can be provided by selecting trainers based on specific requests.
The aggregator class module structure is outlined in Fig. \ref{fig:fig8}, and each module is detailed as follows:
\begin{itemize}
    \item \textbf{Smart contract bridge}: The aggregator uploads the aggregated global model update to IPFS and registers the CID in the smart contract to allow participants to access the global model.
    \item \textbf{Building the global model}: The primary role of the aggregator is to aggregate the local models. They access the smart contract and IPFS to download the list of local models and perform model aggregation according to the specified aggregation algorithm.
    \item \textbf{Aggregation algorithms}: Several aggregation algorithms, including FedAvg, are included in this module. The aggregation algorithms module should be designed to accommodate dynamic decisions of the effective aggregation algorithms for the given task.
    \item \textbf{Client selection algorithms}: After performing aggregation, the aggregator needs to select trainers to participate in the next round. Recently, various client selection algorithms are being actively proposed so that efficient algorithms for the given task in the sense of cost and performance can be chosen. In this study, the option of not performing client selection (all) or excluding trainers based on their performance ranking (scoring order) is provided. In addition to this, there are various client selection algorithms, and we plan to add them so that they can be tested in BCFL as well.
\end{itemize}

\subsubsection{Participant in cross-silo scenario}

\begin{figure*}[!t]
\centering
\includegraphics[width=\linewidth]{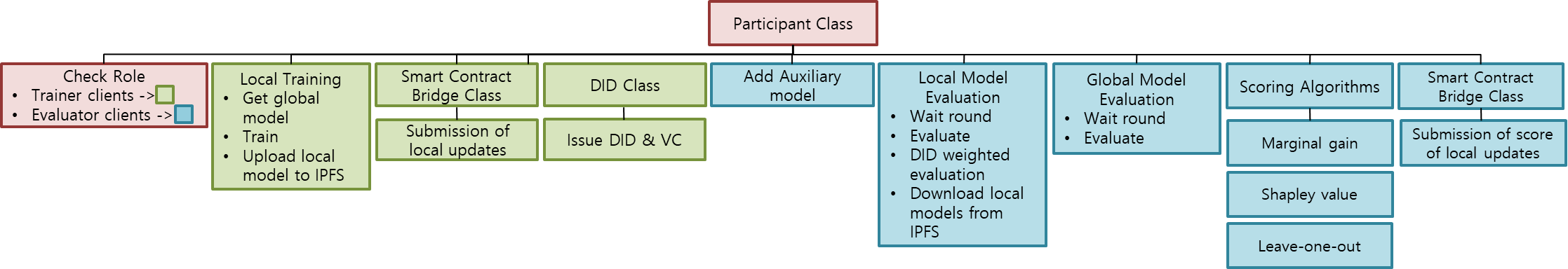}
\caption{Participant class module structure in cross-silo scenario.}
\label{fig:fig9}
\end{figure*}

Fig. \ref{fig:fig9} illustrates the participant module structure in the cross-silo scenario.
In this scenario, there is no dedicated evaluator among the participants.
Instead, multiple auxiliary models are created, with participants alternately assuming the evaluator role.
Thus, each participant in the cross-silo scenario can perform both the roles of an evaluator and a trainer.
The newly considered modules, `check role' and `add auxiliary model,' in Fig. \ref{fig:fig9} are detailed as follows:
\begin{itemize}
\item \textbf{Role check}: In a specific auxiliary model, the creation of the class should be different depending on whether the participant is a trainer or an evaluator.
\item \textbf{Add auxiliary model}: Participants in the evaluator role deploy their local model as the genesis model. Using this genesis model, other participants initiate training based on the cross-device scenario.

\end{itemize}
\subsubsection{BCFL utility functions}

\begin{figure}[!t]
\centering
\includegraphics[width=\linewidth]{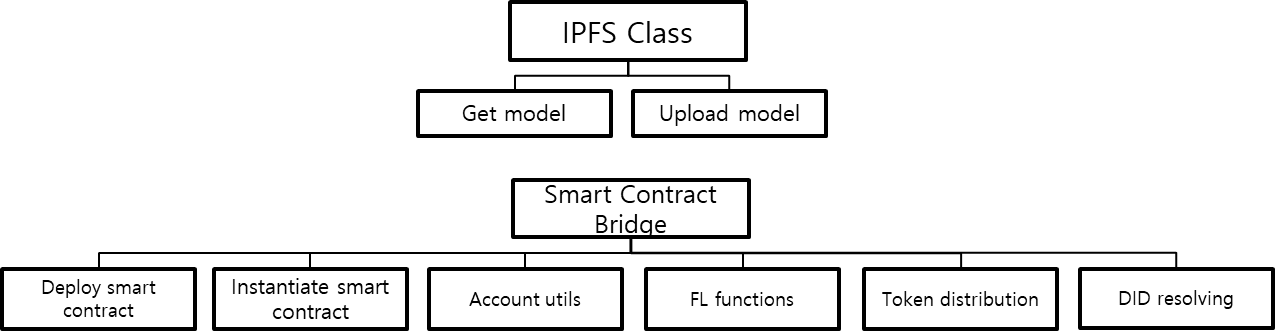}
\caption{BCFL Utilities(IPFS, Smart contract).}
\label{fig:fig10}
\end{figure}

Fig. \ref{fig:fig10} outlines the utility functions required for training.
The IPFS class includes functions necessary for the upload and download of completed local and global models.
As the IPFS connection is established via the HTTP API, a dedicated class is necessary to manage this connection, which is then utilized by all participants.

The smart contract bridge provides contract-related functionalities to all BCFL participants.
This class is implemented using blockchain interaction libraries, such as Web3.py.
It encompasses functionalities like contract deployment, contract instantiation, wallet-related functions (e.g., wallet information retrieval, token deployment, and balance checking), federated learning-related functions, and token-related functions (e.g., token transfer).

\subsection{DID and VC certification system}

\begin{figure}[!t]
\centering
\includegraphics[width=3in]{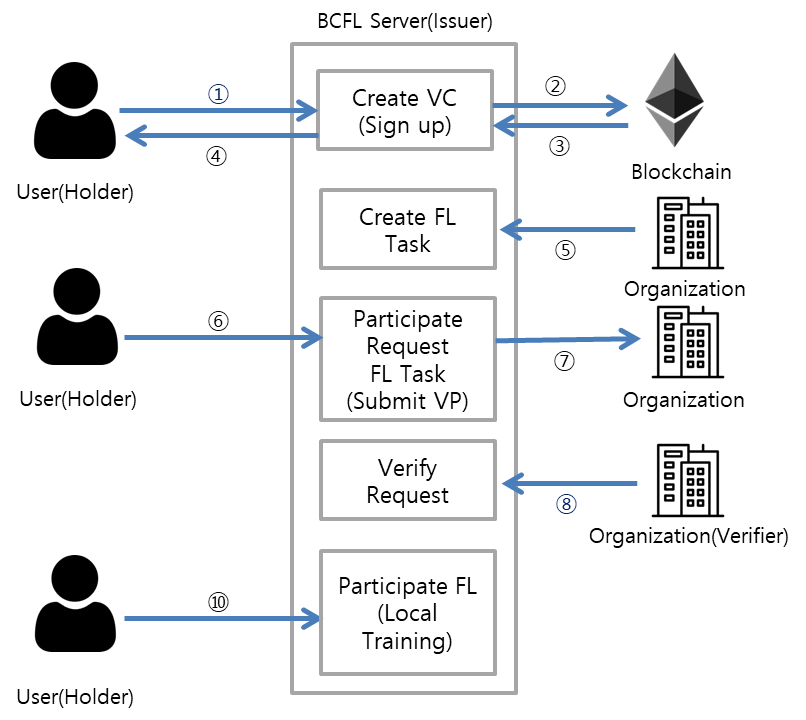}
\caption{DID, VC verifying scenario.}
\label{fig:fig11}
\end{figure}

The authentication system leveraging DID and VC is designed to operate at the service level.
Fig. \ref{fig:fig11} illustrates an example of the system utilization, where the holder refers to a user who engages with the service by issuing a DID.
Upon obtaining the DID, the holder requests VC issuance from the BCFL system and submits it when participating in the federated learning task.
The BCFL system decrypts the VC, verifies whether it is signed by the BCFL system, and sends a verification completion message to the federated learning task registration organization.

\begin{figure*}[!t]
\centering
\includegraphics[width=5in]{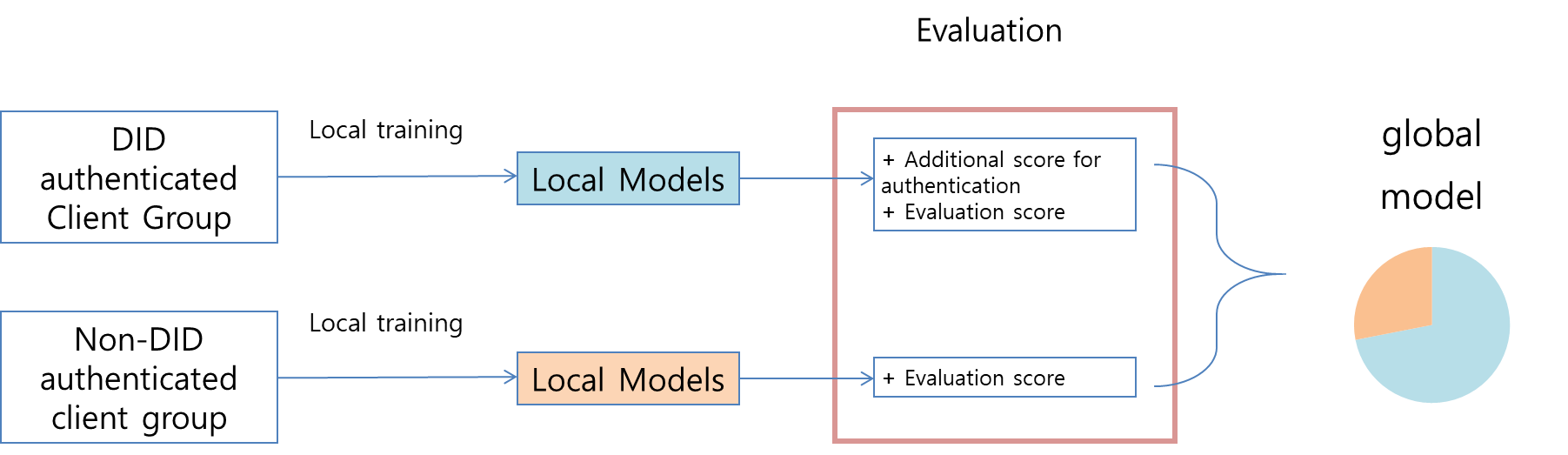}
\caption{Additional score obtained through the DID authentication logic.}
\label{fig:fig12}
\end{figure*}

Fig. \ref{fig:fig12} illustrates the additional score gained through the DID authentication logic.
Users can be divided into authenticated and unauthenticated groups.
For the authenticated group, an additional authentication reward score is granted on top of the evaluation score.
This enables the aggregator, when performing client selection based on the evaluation score, to achieve better performance as the clients are from the authenticated group.
By continuously monitoring and eliminating malicious trainers from the unauthenticated group, it is expected that the performance of the global model will improve.

\section{Experiments}
\label{sec:experiments}

This section presents the experimental results, focusing on our proposed BCFL architecture.
It is important to emphasize that these results integrate aspects of both federated learning and blockchain technologies, rather than examining them separately.
In Section \ref{sec:experiments:sub1}, we discuss the deployment costs of smart contracts in the BCFL framework.
Section \ref{sec:experiments:sub2} investigates the performance of BCFL in non-IID dataset environments, which are prevalent in federated learning scenarios.
Finally, Section \ref{sec:experiments:sub3} critically analyzes the effectiveness of our proposed DID-based authentication system, underscoring its significance and applicability in the context of our research.

\subsection{Gas fee Evaluation}
\label{sec:experiments:sub1}
\begin{table}[!t]
\caption{Smart contract deployment gas price in Sepolia test network.}
\setlength{\tabcolsep}{3pt}
\centering
\begin{tabular}{ll}
\toprule
	Smart contract & Gas price (ETH)\\
\midrule
	Cross-device & $0.005539066784434833$\\
	Cross-silo & $0.013391287720889262$\\
	Token & $0.00293060376676038$\\
	DID-registry & $0.00517602644209044$\\
\midrule
	Total & $0.027474835267814219$ ($51.52$\,USD)\\
\bottomrule
\end{tabular}
\label{tab:table2}
\end{table}

In this work, we verify the reference architecture through deployment and execution in the real-world Ethereum environment.
We utilized the Sepolia test network, an Ethereum test network, to verify the deployment costs of the smart contracts employed in BCFL.
Table \ref{tab:table2} elucidates the deployment costs on the Sepolia test network.
The total deployment cost for the four contracts used in BCFL, including the cross-device scenario, cross-silo scenario, ERC-20 token contract, and DID registry contract, is approximately $0.0274$\,ETH.
This translates to around $51.69$\,USD (or around $66,500$\,KRW) when converted to cash.\footnote{This is based on the exchange rates as of July 22, 2023.}

\begin{table}[!t]
\caption{Smart contract deployment gas price in Ganache local network.}
\setlength{\tabcolsep}{3pt}
\centering
\begin{tabular}{ll}
\toprule
	Samrt contract & Gas price(ETH)\\
\midrule
	Cross-device & $0.04426942$\\
	Cross-silo & $0.1070212$\\
	Token & $0.02332958$\\
	DID-registry & $0.04138968$\\
\midrule
	Total & $0.21600988$ ($403.18$\,USD)\\
\bottomrule
\end{tabular}
\label{tab:table3}
\end{table}

Table \ref{tab:table3} shows the costs of deploying identical contracts on a local Ethereum network via Ganache.
It is noticeable that the deployment costs on the Ganache local network are approximately tenfold that on the Sepolia test network.
The Sepolia test network is configured to mirror the Ethereum mainnet, which adopts the proof of stake (PoS) consensus algorithm.
In contrast, Ganache-configured network emulates the Ethereum 1.0 network that employs the proof of work (PoW) consensus algorithm.
This variance in gas costs arises due to the simulation of differing Ethereum network environments.
Thus, for comprehensive verification from a cost perspective, additional deployment and validation in an environment as close as possible to the main network are necessary.

\subsection{Verification of operation on non-IID dataset}
\label{sec:experiments:sub2}

To verify the operation of BCFL, we first conduct a basic deep learning task on a non-IID dataset.
The experiment, based on the cross-device scenario, utilized FEMNIST dataset for training, which was supplied through LEAF \cite{caldas2018leaf}.
The FEMNIST dataset, comprising $62$ classes of handwritten characters, presents a non-IID distribution for each trainer's data.
The same test dataset was used across all evaluations, and a convolutional neural network (CNN) classification model served as the network.
We set the hyperparameters for training for $15$ global rounds and $2$ local epochs.

\begin{figure}[!t]
\centering
\includegraphics[width=\linewidth]{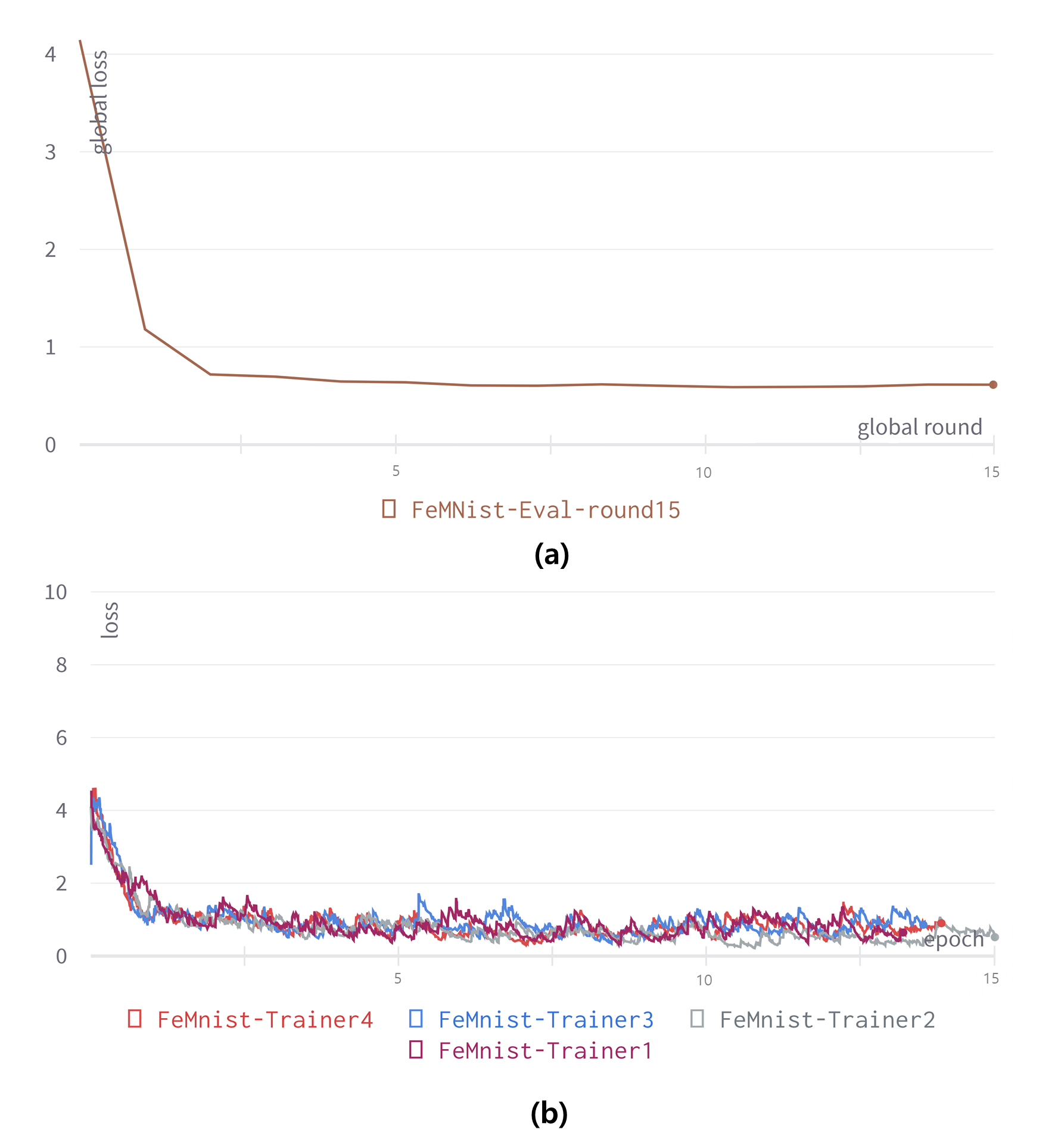}
\caption{Global loss graph (a) and local loss graph (b) for the FEMNIST dataset.}
\label{fig:fig13}
\end{figure}

Fig. \ref{fig:fig13} shows the progression of global and local loss as four trainers participate in training based on the FEMNIST dataset.
In terms of global loss, as depicted in Fig. \ref{fig:fig13}(a), it initiated at $4.15$ in the first round and gradually converged to approximately $0.61$ by the final round.
Concurrently, as depicted in Fig. \ref{fig:fig13}(b), the local loss experienced by the trainers also displays a converging pattern over time.
As the $15$ global rounds progress, the local model loss of all four trainers shows similar signs of convergence.
This demonstrates the effectiveness of the training procedure and its ability to minimize loss across all participating trainers.

%the graph of global loss and local loss when four trainers participated in training with the FEMNIST dataset. For the global loss (a), it started at 4.15 in the first round but converged to approximately 0.61 in the final round. The local loss (b) according to cumulative epochs showed a convergence trend as the 15 global rounds progressed, with the local model loss of the four trainers converging as well.

\subsection{Global model with DID authentication system}
\label{sec:experiments:sub3}

\begin{figure}[!t]
\centering
\includegraphics[width=\linewidth]{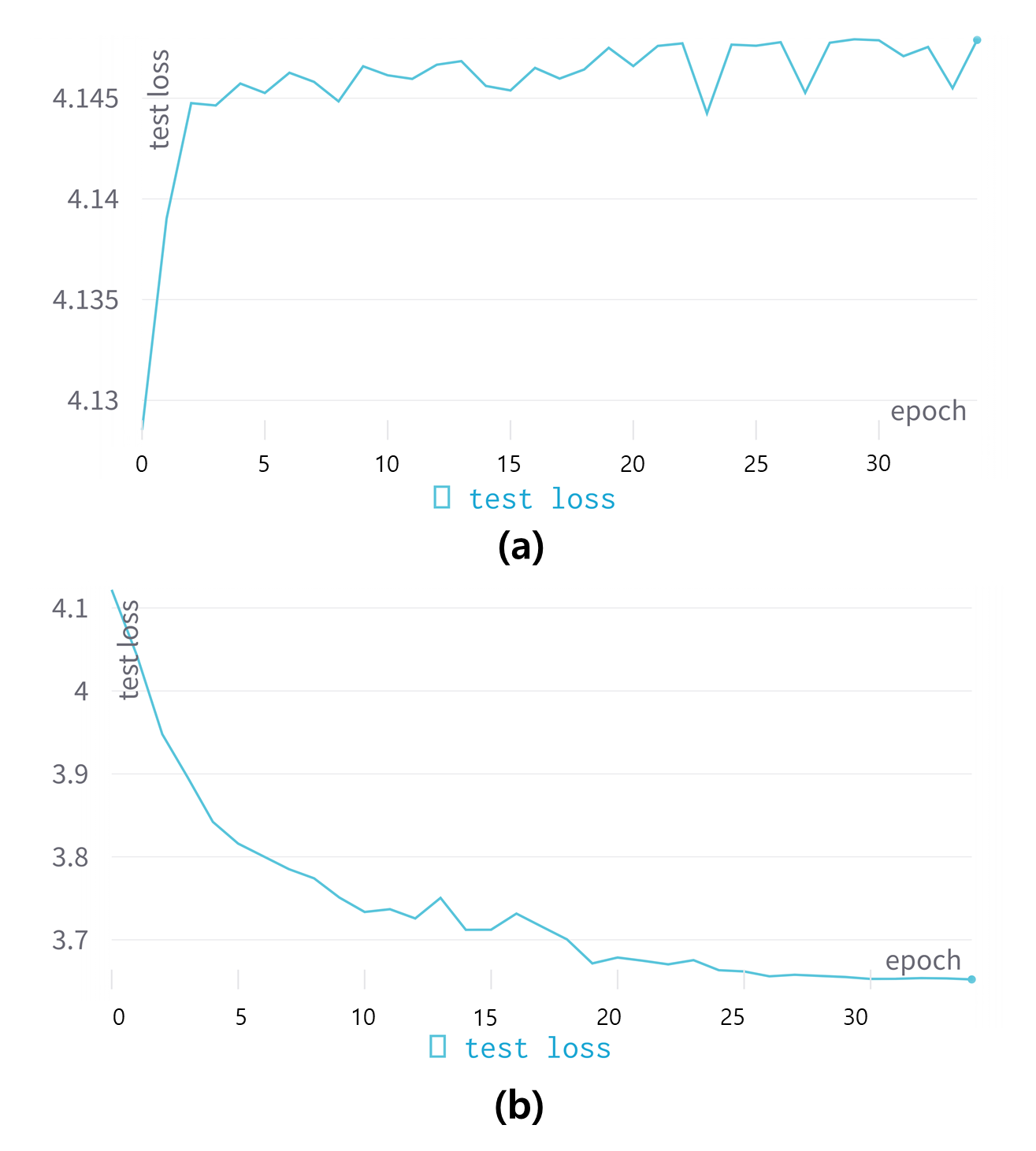}
\caption{Test loss graph of DID non-certified trainer (a) and test loss graph of DID certified trainer (b).}
\label{fig:fig14}
\end{figure}

In this experiment, we seek to show an example scenario of the integrated architecture of BCFL and DID systems.
For this purpose, we set up a total of $25$ trainers, each with a unique wallet address on the Ganache network.
All trainers maintain their own training dataset in their local environments, utilizing the FEMNIST dataset.
Intentionally, we set $12$ of these $25$ trainers to have normal datasets, whereas the remaining $13$ to have label-flipped datasets.
These $13$ trainers are treated as malicious trainers, and thus, we have them not participate in the DID authentication process.
Fig. \ref{fig:fig14} shows the test loss graph resulting from evaluations conducted on these trainers.
More specifically, Fig. \ref{fig:fig14}(a) represents the test loss from the unauthenticated trainers, demonstrating poor performance.
This result implies that their aggregation into the global model could potentially exert a detrimental impact.
In contrast, Fig. \ref{fig:fig14}(b) represents the test loss from trainers who have undergone DID authentication and have normal datasets, showing a trend of decreasing loss as training progresses.
The results suggest that the DID authentication process effectively separates trainers with normal datasets from those with label-flipped datasets, contributing to the improvement of the overall model performance.

%For this purpose of training, a total of $25$ trainers are set up, each possessing a unique wallet address on the Ganache network.
%All trainers maintain their own training dataset in their local environments, using the FEMNIST dataset.
%We have intentionally arranged for the $25$ trainers such that $12$ of them have normal datasets, while the remaining $13$ possess label-flipped datasets.
%These $13$ trainers are treated as malicious trainers, and accordingly, they do not undergo the authentication process using DID.
%Fig. \ref{fig:fig14} shows the results of the test loss graph from the evaluations conducted on these trainers.
%Specifically, Fig. \ref{fig:fig14}(a) shows the test loss from the malicious trainers, while Fig. \ref{fig:fig14}(b) shows the test loss from the non-malicious trainers.
%According to the figures, malicious trainers demonstrate inferior performance.
%This is because they have not been certified.
%This suggests that their aggregation into the global model may exert a detrimental impact.
%On the other hand, the non-malicious trainers have a decreasing trend as the training progresses, i.e., good performance, because they have undergone DID authentication and possess normal datasets.

\begin{figure}[!t]
\centering
\includegraphics[width=\linewidth]{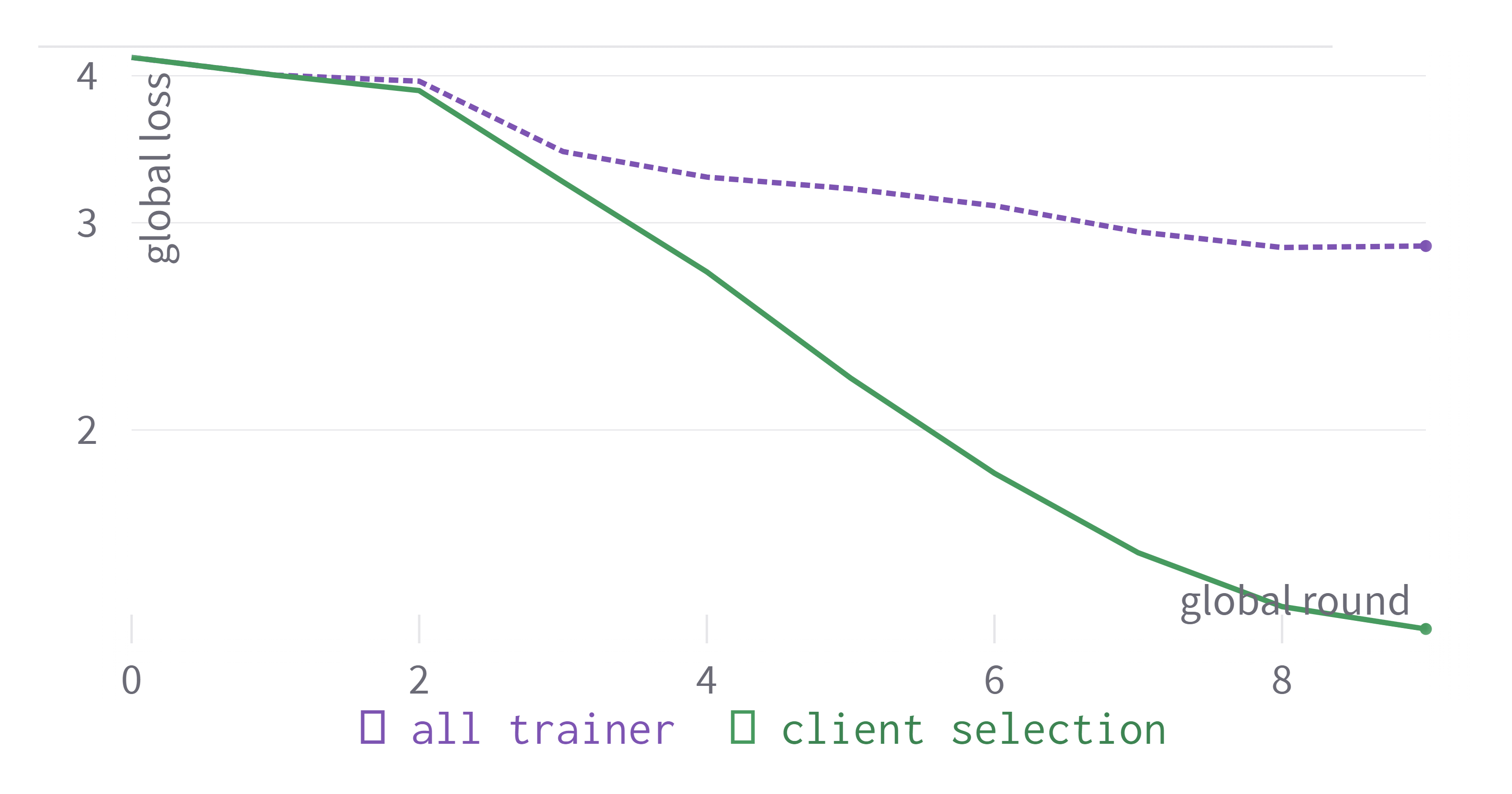}
\caption{BCFL's global model loss graph where client selection was performed by giving additional scores to DID-certified trainers(solid) and BCFL's global model loss graph with no client selection(dashed)}
\label{fig:fig15}
\end{figure}

Next, to assess the potential impact of the persistent participation of unauthenticated malicious trainers in training, we examine the performance of the global model under a different scenario.
In this context, regular trainers that undergo DID authentication are rewarded with an extra score, which is equivalent to $10$\,\% of the score achieved from the local evaluation.
As shown in Fig. \ref{fig:fig15}, when we grant additional scores to DID-authenticated clients (depicted in green) and juxtapose this with a scenario in which no extra scores are granted (depicted in purple), a substantial decrease in global loss is observed, dropping from approximately $2.9$ to $1.3$.
This result underscores the effectiveness of rewarding DID-authenticated trainers in reducing global loss.

The experimental results presented in this section demonstrate the effective integration of BCFL with DID.
The detailed evaluation of training loss with different sets of trainers validates the robustness and versatility of our proposed system.
Moreover, the added security and accountability from the DID authentication system show its potential in mitigating the risks associated with malicious trainers.
As our architecture continues to be refined and improved, it can further accommodate the evolving needs of federated learning environments.
We anticipate these results to contribute significantly to the development of secure, efficient, and practical BCFL systems.

\section{Conclusions and Future Directions}
\label{sec:conclusion}
BCFL presents a promising decentralized solution, merging the advantages of federated learning and blockchain technologies.
This study introduces a novel reference architecture for BCFL with a thorough analysis of the key components and processes.
Furthermore, we implemented and verified the architecture in a practical, real-world Ethereum development environment.

For the verification of the architecture, we first contrasted the deployment costs of smart contracts on the Ethereum Sepolia test network, which is closer to the real-world main network environment, with those on the Ethereum local network simulated using Ganache.
The experimental results show considerable discrepancies in deployment costs between real production and development environments.
For future work, more intensive testing and verification, including operational cost and transaction processing speed for an improved user experience, are required.
Second, to verify the FL process running on Ethereum simulation networks constructed using Ganache, we demonstrated convergence trends for both global and local models in terms of loss.
Third, for verification and as a prominent example of extensibility, DIDs have been successfully integrated as an authentication method to introduce practical utilization within BCFL. When an additional score is awarded to authenticated trainers during client selection for global model evaluation, a notable performance enhancement is observed, verifying the potential value of incorporating DID authentication systems within BCFL in a real-world deployment. We conceptualized and conducted experiments to demonstrate the effective functioning of this authentication system. While the application of an authentication system to BCFL for managing and tracking malicious trainers is promising, further research is warranted to determine the legitimacy of adopting DID as the authentication mechanism.

% esgoh : Reviewer 1, Concern 4
This study primarily focuses on the practical verification of the architecture's functionality, while future research will shift towards evaluating the system's performance, taking into account performance indicators such as the accuracy of federated learning models and the evaluation of contributions.
The fair and efficient assessment of participating clients' contributions is particularly critical in BCFL, as it directly influences the model accuracy and the incentive mechanism.
To advance our proposed approach, it is essential to integrate evolving client evaluation techniques from the federated learning domain, thus qualitatively improving global models.
Moreover, expanding the range of client selection methods may also be advantageous.
Fundamentally, the continued integration of cutting-edge techniques in federated learning is vital, requiring that task generators be supported in selecting and formulating BCFL tasks appropriately.

In addition, operational verifications and performance evaluations will take place on an actual Ethereum main network or an equivalent test network. Possibilities such as utilizing Ethereum client programs to establish and operate a gas-free private network will also be investigated. Alternatively, commercially available main networks with rapid transaction processing speeds and low gas fees could serve as suitable platforms for the real deployment of BCFL. A Layer-2 Architecture may be a promising solution for this approach.
Building upon these explorations, our future work will also include a detailed analysis of communication costs within the BCFL system.
This will particularly focus on the interaction between IPFS and blockchain, aiming to further enhance the architecture's efficiency and applicability.

\bibliographystyle{IEEEtran}
\bibliography{IEEEabrv,BCFL_nohighlight}

\begin{IEEEbiography}[{\includegraphics[width=1in,height=1.25in,clip,keepaspectratio]{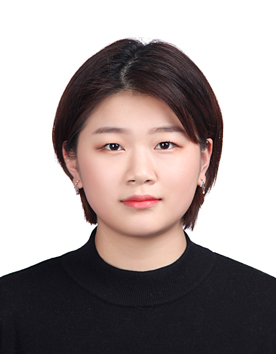}}]{Eunsu Goh}
received the B.S. degree in electronics and communications engineering from Kwangwoon University, Seoul, South Korea, in 2019.
She has been immersed in the realm of deep learning.
In 2021, she received the M.S. degree from the same institution.
Her interest gravitates towards algorithms from federated learning and blockchain technology, particularly in creating trustworthy collaborative learning systems. 
\end{IEEEbiography}

\begin{IEEEbiography}[{\includegraphics[width=1in,height=1.25in,clip,keepaspectratio]{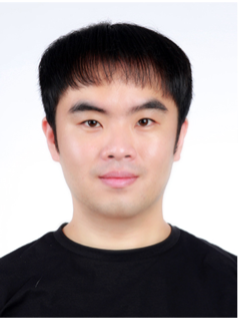}}]{Dae-Yeol Kim}
received the B.S. degree in electronics and communications engineering from Kwangwoon University, Seoul, South Korea, in 2016, and the Ph.D degree in electronics and communications engineering from Kwangwoon University, Seoul, South Korea, in 2022.
From 2016 to 2019, he served an Associate Research Engineer at TvStorm, Seoul, South Korea.
Since 2022, he assumed the role of Senior Research Engineer at InnopiaTech, Sungnam-si, Korea.
His research interests include medical artificial intelligence, Computer vision, and Blockchain-enabled Federated Learning.
\end{IEEEbiography}

\begin{IEEEbiography}[{\includegraphics[width=1in,height=1.25in,clip,keepaspectratio]{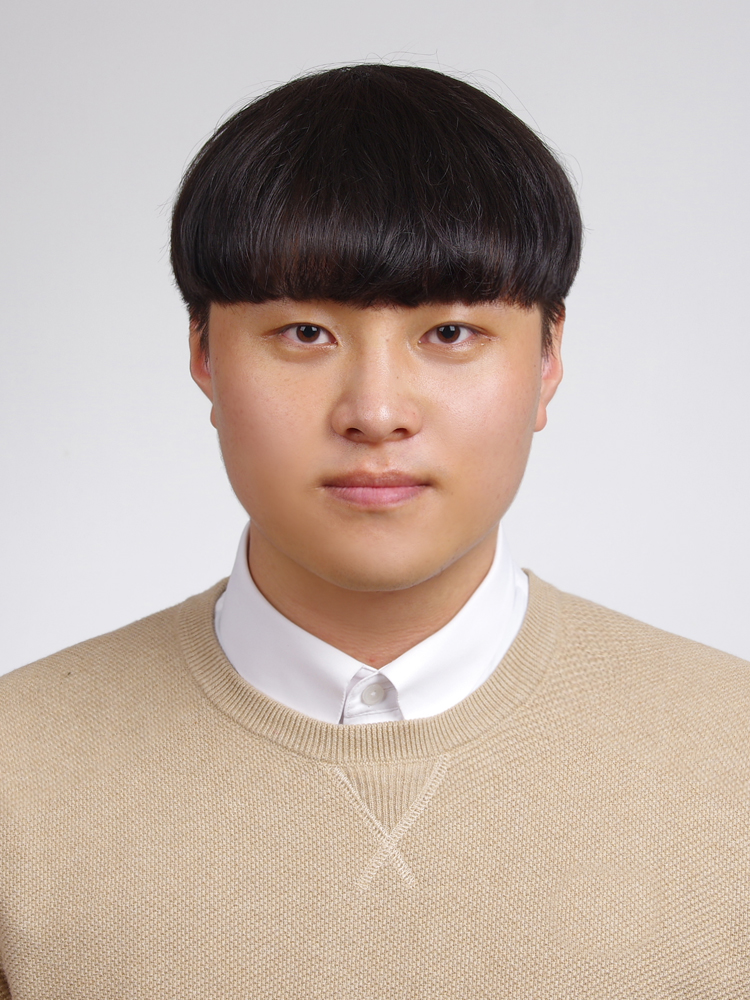}}]{Suyeong Oh}
B.S in Kwangwoon University, Department of Electronics and Communications Engineering.
M.E in Kwangwoon University, Department of Electronics and Communications Engineering.
2021~2023. Associate Research Engineer in Tvstorm to develop Android TV application.
His research interests in developing integrated platform with Artificial Intelligence and Block Chain.
\end{IEEEbiography}
\vfill
\newpage

\begin{IEEEbiography}[{\includegraphics[width=1in,height=1.25in,clip,keepaspectratio]{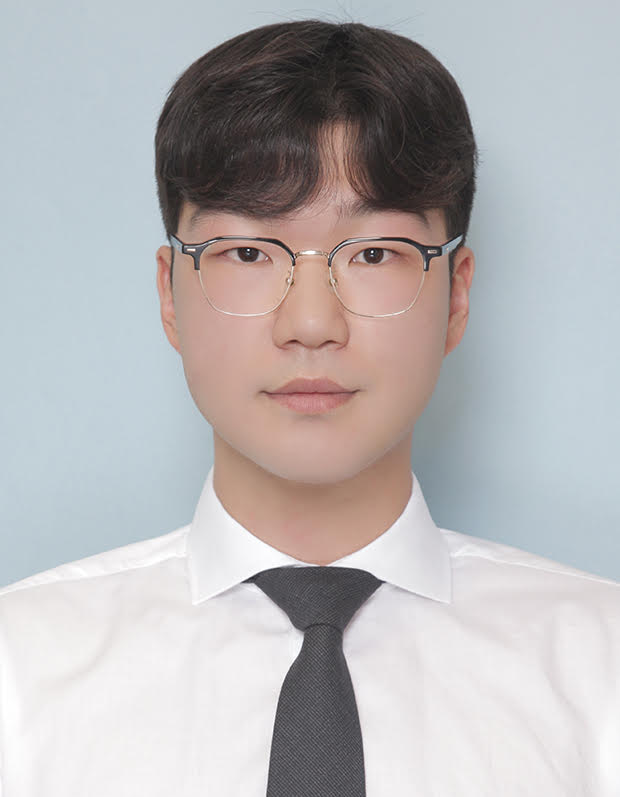}}]{Jong-Eui Chae}
received a B.S. degree in electronics and communications engineering in 2022, and he is currently pursuing a M.S. degree from Kwangwoon University, Seoul, South Korea.
His research interests include vital-signal engineering, data analytics through deep learning.
\end{IEEEbiography}

\begin{IEEEbiography}[{\includegraphics[width=1in,height=1.25in,clip,keepaspectratio]{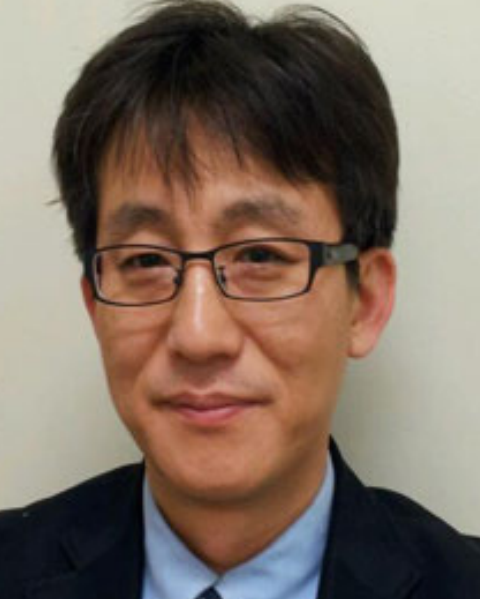}}]{Kwangkee Lee}
received the B.S., M.S., and Ph.D. degrees in electronics engineering from Yonsei University, Seoul, South Korea, in 1986, 1988, and 1993, respectively.

From 1994 to 2014, he was a Researcher with the Samsung Advanced Institute of Technology and Samsung Electronics.
From 2016 to 2019, he was an Industrial Convergence PD for R\&BD planning with the Ministry of Industry, Industrial Technology Evaluation and Management Institute.
He is currently a software architect \& principal investigator with Innopia Technologies Inc.
\end{IEEEbiography}

\begin{IEEEbiography}[{\includegraphics[width=1in,height=1.25in,clip,keepaspectratio]{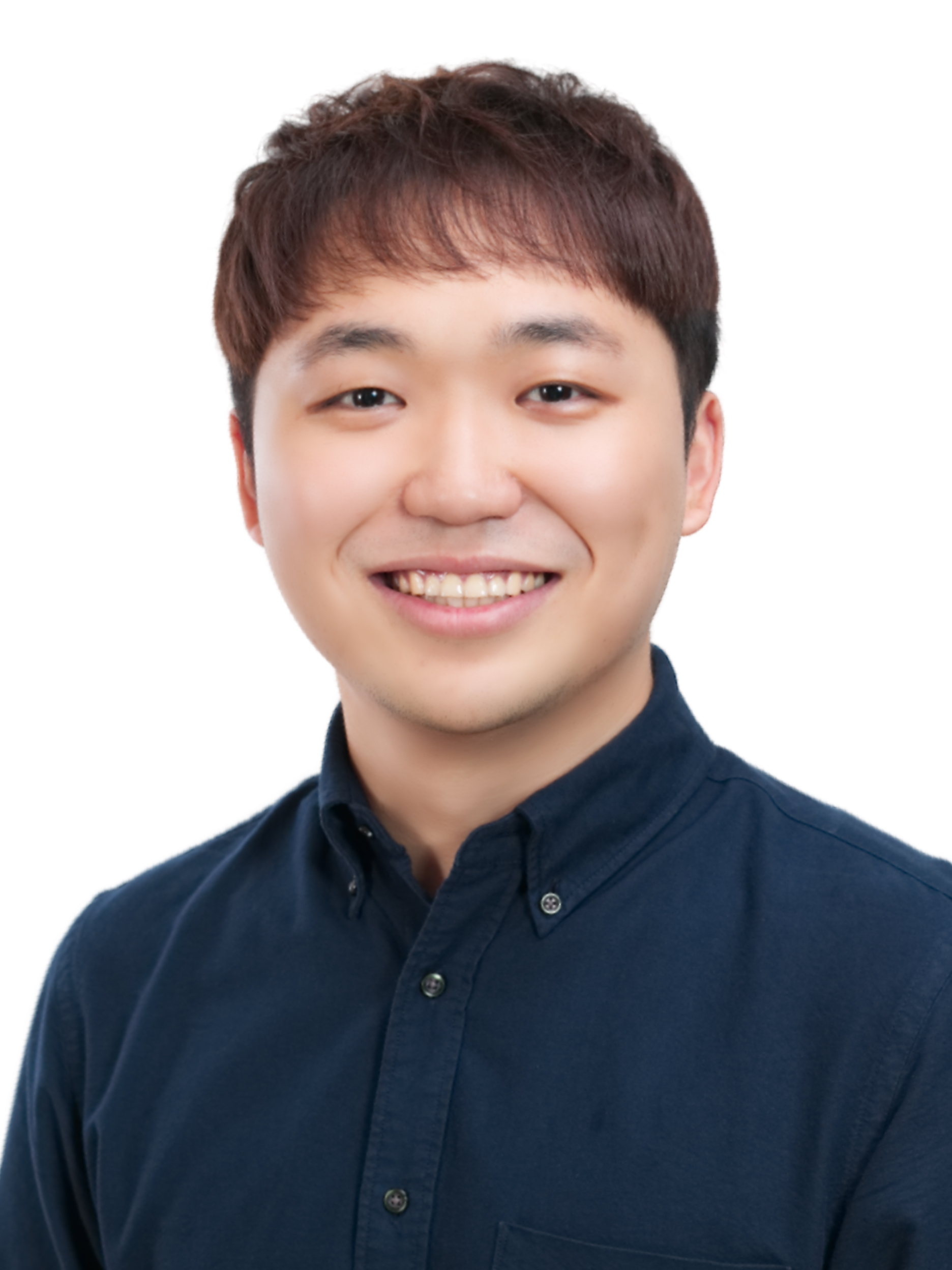}}]{Do-Yup Kim}
(Member, IEEE) received his B.S. degree \textit{(summa cum laude)} in Electronics and Communications Engineering from Kwangwoon University, Seoul, South Korea, in 2016, and his Ph.D. in Electrical and Electronic Engineering from Yonsei University, Seoul, South Korea, in 2022.

From 2021 to 2022, he was a Visiting Scholar with the Department of Electrical and Computer Engineering, Virginia Tech Research Center, Arlington, VA, USA, and then a Post-Doctoral Scholar with the Department of Electrical and Computer Engineering, The Ohio State University, Columbus, OH, USA.
Since September 2022, he has been an Assistant Professor with the Department of Information and Communication AI Engineering, Kyungnam University, Changwon-si, Gyeongsangnam-do, South Korea.
His research interests include communication networks, optimization, and machine learning.
\end{IEEEbiography}
\vfill

\EOD

\end{document}